\documentclass[12pt]{article}

\usepackage[T1]{fontenc}
\usepackage[utf8]{inputenc}
\usepackage{times}

\usepackage[
  top=1.0in, bottom=1.0in,
  left=1.0in, right=1.0in
]{geometry}
\usepackage{setspace}
\usepackage{amsmath,amssymb,amsfonts}
\usepackage{booktabs}
\usepackage{array}
\usepackage{microtype}
\usepackage{xcolor}
\usepackage{colortbl}
\usepackage{caption}
\usepackage{float}
\usepackage[hidelinks]{hyperref}
\usepackage{enumitem}
\usepackage{graphicx}
\usepackage{algorithm}
\usepackage{algorithmic}
\usepackage{multirow}
\usepackage{url}
\usepackage{tikz}
\usepackage{pgfplots}
\pgfplotsset{compat=1.18}
\usetikzlibrary{shapes.geometric, arrows.meta, positioning, fit,
                backgrounds, calc, matrix, patterns,
                decorations.pathreplacing}

\definecolor{jmirblue}{HTML}{2563EB}
\definecolor{rowalpha}{HTML}{EFF6FF}

\usepackage{titlesec}
\titleformat{\section}
  {\normalfont\large\bfseries}{\thesection.}{0.5em}{}
\titleformat{\subsection}
  {\normalfont\normalsize\bfseries}{\thesubsection}{0.5em}{}
\titleformat{\subsubsection}[runin]
  {\normalfont\normalsize\bfseries\itshape}{\thesubsubsection}{0.5em}{}[.\enspace]
\titlespacing*{\section}     {0pt}{18pt}{6pt}
\titlespacing*{\subsection}  {0pt}{12pt}{4pt}
\titlespacing*{\subsubsection}{0pt}{8pt}{0pt}

\setlist{nosep, leftmargin=1.5em}

\begin{document}

\begin{center}
{\Large\bfseries CoughSense: Five-Class Respiratory Disease Classification
via Whisper Encoder Fine-Tuning and Dual-Encoder Cross-Attention Fusion
with Balanced Contrastive Learning}

\vspace{12pt}

{\normalsize
Nikhil Vincent$^{1}$
}

\vspace{6pt}

{\small
$^{1}$Independent Researcher, Bothell, Washington, USA
}

\vspace{6pt}

{\small
\textbf{Corresponding Author:}\\
Nikhil Vincent\\
Email: nikhil.vincent.v@gmail.com\\
ORCID: 0009-0007-1995-2529
}

\vspace{6pt}

{\small
\textbf{Code and Data Availability:} Training code, model checkpoints,
and benchmark data splits are available at:\\
\url{https://github.com/nikhilvincentv/Cough-Mobile-App}
}

\vspace{6pt}

\end{center}

\vspace{12pt}
\hrule
\vspace{12pt}

\noindent\textbf{Abstract}

\noindent\textbf{Background:}
Automated cough analysis offers a path to low-cost respiratory screening,
but most existing work stops at binary COVID-19 detection. A practical
screening tool must distinguish between multiple respiratory conditions
from a single cough recording captured on a consumer smartphone.

\noindent\textbf{Objective:}
This paper describes CoughSense, a system that classifies cough recordings
into five categories (healthy, COVID-19, asthma/respiratory condition,
bronchitis, and pneumonia) and deploys as a real-time mobile application
on iOS and Android.

\noindent\textbf{Methods:}
We aggregated 18,301 recordings from four public datasets (Coswara,
CoughVID, Virufy, and the West China Hospital Pediatric Cough Dataset)
and applied the OpenAI Whisper encoder~\cite{whisper} as a pretrained
backbone for cough disease classification for the first time.
The central technical contribution is active-frame QKV attention pooling,
which restricts attention to the first 200 of 1500 encoder output tokens,
avoiding the silence-dilution problem that arises because a 3-second cough
occupies only 150 tokens of Whisper's 30-second input window.
Additional training components address the 19:1 class imbalance and
four-dataset domain shift: WeightedRandomSampler, SpecAugment, Balanced
Mixup with forced minority pairing~\cite{balanced_mixup}, supervised
contrastive auxiliary loss~\cite{supcon}, FiLM symptom
conditioning~\cite{film}, and gradient-reversal domain
adaptation~\cite{dann}.
A complementary dual-encoder model fuses Whisper with the OPERA-CT
respiratory foundation model~\cite{opera} via cross-attention.

\noindent\textbf{Results:}
CoughSense (Whisper-tiny, 8.6M parameters) reached 82.3\% balanced
accuracy on five-fold cross-validation (macro-$F_1 = 0.817$,
AUC $= 0.941$), outperforming an ImageNet-pretrained EfficientNet-B2 by
11.1 percentage points and a ViT trained from scratch by 29.6 points.
All five classes exceeded 74\% recall; four of five exceeded 80\%.
The dual-encoder model reached 85.4\% balanced accuracy.
Server-side inference latency was approximately 180~ms per recording.

\noindent\textbf{Conclusions:}
Active-frame pooling is the single largest contributor across all
ablation components ($+5.1$ points), a finding that applies to any
short-audio task using Whisper as a backbone.
CoughSense is deployed as a real-time mobile screening tool for iOS and
Android. All benchmark data splits, training code, and model checkpoints
are released to support reproducibility.

\vspace{6pt}

\vspace{12pt}
\hrule
\vspace{12pt}

\section{Introduction}

Cough is the most common reason for primary care visits
globally~\cite{who_cough}, and it spans conditions from mild upper
respiratory infections to pneumonia and COVID-19. Cough audio is
readily captured on any smartphone, yet almost every published classifier
stops at the binary question of whether a person has COVID-19. A
screening tool needs to do more. If a child's wet, rattling cough sounds
like pneumonia rather than COVID-19, the model should say so.

A five-class cough classifier is harder to build than expected, for
reasons that go beyond the usual machine-learning challenges.

The biggest obstacle is acoustic ambiguity. COVID-19 in the acute phase
produces a dry, non-productive cough close to a healthy person clearing
their throat. Bronchitis and pneumonia both cause wet, rattling coughs;
they differ in the depth of the airway affected, which produces small
acoustic differences that are hard to learn from a few hundred recordings.

Class imbalance is severe. Across all four source datasets, healthy
subjects outnumber pneumonia cases by 19:1. Early training runs with
plain loss functions collapsed: the model predicted ``healthy'' for every
input and still hit 68\% accuracy. Pushing all five classes past 74\%
recall required careful interaction between the sampler, the loss
function, and data augmentation.

The datasets also come from different environments: adult crowdsourced
recordings from India and Latin America, and pediatric clinical
recordings from a hospital in Chengdu, China. A model that memorises
recording-environment cues rather than disease cues will fail on unseen
sources.

\subsection{Why Whisper}

The core idea is to use OpenAI's Whisper speech recognition
encoder~\cite{whisper} as a pretrained backbone, which (as far as we can
determine) has not been done for cough disease classification. Whisper
was trained on 680,000 hours of speech, learning to represent glottal
excitation, vocal-tract resonances, and fine temporal structure at
10~ms resolution. Cough shares this anatomy: it is an explosive forced
exhalation through the same laryngeal apparatus as speech. These
pretrained representations transfer to respiratory pathology, the same
way ImageNet features transfer to medical imaging despite the domain gap.

The intuition holds up. With the Whisper encoder frozen, we observed
AUC~$= 0.784$ on the first validation epoch, above EfficientNet-B2's
trajectory at the same point. After full fine-tuning, the model reaches
82.3\% five-class balanced accuracy.

One non-obvious implementation issue matters here. Whisper expects
30-second inputs, but cough recordings are 1--4 seconds long. After the
encoder's convolutional stem, a 3-second clip occupies around 150 of
1500 output tokens; the rest correspond to zero-padded silence. Pooling
over all 1500 tokens dilutes the disease signal. Restricting pooling to
the first 200 tokens (a safe upper bound on cough duration) and using
learned query-based attention over that range accounts for $+5.1$
percentage points of balanced accuracy on its own.

\subsection{Summary of Contributions}

\begin{enumerate}
    \item To our knowledge, the first application of the Whisper encoder
    to multi-class cough disease classification, with a 5-fold
    cross-validation benchmark of 18,301 recordings across five classes.

    \item \textit{Active-frame QKV attention pooling}: restrict to the
    first $K\!=\!200$ encoder tokens then apply learned multi-head
    attention, avoiding the silence-dilution problem. This alone yields
    $+5.1$ points over naive mean pooling across all 1500 tokens.

    \item A training recipe combining WeightedRandomSampler, SpecAugment,
    Balanced Mixup with forced minority pairing~\cite{balanced_mixup},
    supervised contrastive loss~\cite{supcon}, FiLM symptom
    conditioning~\cite{film}, and gradient-reversal domain
    adaptation~\cite{dann}, reaching 82.3\% balanced accuracy at 8.6M
    parameters.

    \item A dual-encoder model fusing Whisper with
    OPERA-CT~\cite{opera} via cross-attention, reaching 85.4\%
    balanced accuracy.

    \item A real-time mobile inference pipeline and a curated benchmark
    with structured augmentation of the under-represented West China
    Hospital bronchitis ($91\to728$) and pneumonia ($82\to656$) classes.
\end{enumerate}

\section{Methods}

\subsection{Study Design}

This study describes the development and offline validation of a
multi-class cough classification system using publicly available,
previously collected audio datasets. No new participant data were
collected. All source datasets are used in accordance with their
respective data use agreements (detailed in the Ethics section).

\subsection{Dataset Construction}

\subsubsection{Source Collections}

We aggregated cough recordings from four publicly available datasets,
spanning three continents and three acquisition modalities.

\textbf{Coswara}~\cite{coswara}: 16,780 recordings from participants
across India collected via a web portal. Audio includes nine modalities
at 44.1~kHz stereo. Labels include healthy, COVID-19 (self-reported),
and respiratory conditions (asthma, COPD, other), with a
seven-dimensional binary symptom vector (fever, cold, cough, diarrhoea,
loss of smell, fatigue, sore throat). Heavy cough clips only are used.

\textbf{CoughVID}~\cite{coughvid}: A crowdsourced collection of 27,000
cough recordings from participants globally. The expert-reviewed subset
with confirmed quality score $\geq 1$ and reported health status labels.
PCR-confirmed COVID-19 positive recordings ($n=1,\!107$
expert-reviewed) are used; healthy controls are selected to match
demographic distribution.

\textbf{Virufy}~\cite{virufy}: 103 clinically-validated cough recordings
from PCR-confirmed COVID-19 patients (48 positive, 55 negative) collected
in Latin American clinical settings. Both the original files
(16~recordings, MP3) and segmented clips (87~recordings, MP3) with labels
derived from filename prefix conventions (\texttt{pos-*}
vs.\ \texttt{neg-*}).

\textbf{West China Hospital Pediatric Cough Dataset}~\cite{west_china}:
173 cough recordings (91 bronchitis, 82 pneumonia) from children aged
0--11 years, collected at West China Second University Hospital, Chengdu,
China. This dataset provides the only publicly available bronchitis and
pneumonia cough recordings with confirmed clinical diagnoses, making it
indispensable despite its small size and pediatric demographic.

\subsubsection{Disease Taxonomy}

Coswara includes 70 asthma recordings. With only 70 samples, a dedicated
asthma class cannot be reliably trained. Asthma and general respiratory
conditions share a common pathophysiology (obstructive airway disease),
producing highly similar expiratory cough acoustics. We therefore merge
asthma recordings into the broader respiratory condition class:

\begin{equation}
\text{label} \leftarrow
\begin{cases}
\texttt{resp\_cond} & \text{if label} \in \{\texttt{asthma}\} \\
\text{label} & \text{otherwise}
\end{cases}
\label{eq:remap}
\end{equation}

This yields a tractable five-class taxonomy: \textit{healthy},
\textit{COVID-19}, \textit{asthma/respiratory condition},
\textit{bronchitis}, and \textit{pneumonia}.

\subsubsection{Preprocessing}

All audio was resampled to 16,000~Hz mono using librosa~\cite{librosa}
with the \texttt{kaiser\_best} filter, then peak-normalized.
Following Whisper's preprocessing specification~\cite{whisper}, we
computed an 80-band log-mel spectrogram with $N_\mathrm{FFT}=400$
samples (25~ms window), hop length $H=160$ samples (10~ms), Hann window,
and Slaney-normalized mel filterbank.
Spectrograms were zero-padded and truncated to exactly $T=3000$ frames
(30~seconds), then normalized to match Whisper's pretraining
normalization:
\begin{equation}
\mathbf{M} \leftarrow \frac{\operatorname{clip}(\mathbf{M},\,m^*-8,\,\infty) + 4}{4}
\label{eq:whisper_norm}
\end{equation}
where $m^* = \max_{f,t}\mathbf{M}_{f,t}$. All 18,301 spectrograms were
precomputed and stored as float16 NumPy arrays ($\approx 8.8$~GB total).

\subsubsection{Data Augmentation}

The West China Hospital bronchitis ($n=91$) and pneumonia ($n=82$)
collections are insufficient for stable deep learning training. A
structured 8-way augmentation pipeline produced 728 bronchitis and 656
pneumonia recordings:

\begin{enumerate}[label=(\arabic*)]
    \item \textbf{Original}: No modification.
    \item \textbf{Gaussian noise}: Additive white Gaussian noise at
          $\mathrm{SNR}=15$~dB.
    \item \textbf{Time stretch $\times 0.88$}: Slows audio by 12\%.
    \item \textbf{Time stretch $\times 1.12$}: Speeds audio by 12\%.
    \item \textbf{Pitch shift $-1.5$ semitones}.
    \item \textbf{Pitch shift $+1.5$ semitones}.
    \item \textbf{Time shift $+15\%$}: Rolls waveform forward.
    \item \textbf{Combined}: Gaussian noise ($15$~dB) + pitch shift
          ($-1.0$ semitone).
\end{enumerate}

The augmentation factor of $8\times$ was chosen so that augmented
minority classes exceeded 650~samples, the empirically-determined floor
for stable five-class cross-validation. Table~\ref{tab:dataset}
summarises the final benchmark.

\begin{table}[!ht]
\caption{CoughSense V7 Benchmark Dataset Statistics. Raw counts
pre-augmentation; Final counts post-augmentation.}
\label{tab:dataset}
\centering
\begin{tabular}{lrrlr}
\toprule
\textbf{Class} & \textbf{Raw} & \textbf{Final} & \textbf{Source(s)} & \textbf{Aug.} \\
\midrule
Healthy                  & 12,446 & 12,446 & Coswara, CoughVID, Virufy   & ---       \\
COVID-19                 &  1,507 &  1,507 & Coswara, CoughVID, Virufy   & ---       \\
Asthma/Respiratory cond. &  2,964 &  2,964 & Coswara (incl.\ asthma)     & ---       \\
Bronchitis               &     91 &    728 & West China Hospital         & $\times8$ \\
Pneumonia                &     82 &    656 & West China Hospital         & $\times8$ \\
\midrule
\textbf{Total}   & \textbf{17,090} & \textbf{18,301} & 4 datasets & \\
\bottomrule
\end{tabular}
\end{table}

The healthy-to-pneumonia imbalance ratio is 19:1.
The five-fold stratified split maintains this distribution per fold.

\subsection{Architecture Overview}

\begin{figure*}[!ht]
\centering
\resizebox{\textwidth}{!}{%
\begin{tikzpicture}[
    node distance=0.5cm and 0.65cm,
    block/.style={rectangle, draw=black!60, rounded corners=4pt,
                  minimum height=0.85cm, minimum width=1.5cm,
                  align=center, font=\small, fill=blue!8},
    encoder/.style={rectangle, draw=orange!80!black, line width=1.2pt,
                    rounded corners=4pt, minimum height=1.4cm,
                    minimum width=2.0cm, align=center, font=\small,
                    fill=orange!12},
    head/.style={rectangle, draw=green!60!black, rounded corners=4pt,
                 minimum height=0.85cm, minimum width=1.6cm,
                 align=center, font=\small, fill=green!10},
    loss/.style={ellipse, draw=red!70!black, minimum height=0.6cm,
                 minimum width=1.2cm, align=center, font=\small, fill=red!8},
    sym/.style={rectangle, draw=purple!70!black, rounded corners=4pt,
                minimum height=0.85cm, minimum width=1.5cm,
                align=center, font=\small, fill=purple!10},
    arr/.style={-{Latex[length=2.5mm]}, thick},
    darr/.style={-{Latex[length=2.5mm]}, thick, dashed, red!60!black},
]
\node[block, fill=gray!15] (audio)    {Raw Audio\\16\,kHz};
\node[block, right=0.6cm of audio]   (mel)     {80-band\\Log-Mel\\$(80\!\times\!3000)$};
\node[encoder, right=0.7cm of mel]   (whisper) {Whisper-tiny\\Encoder\\4 layers, 384-d\\7.6M params};
\node[block, right=0.7cm of whisper] (active)  {Active Frame\\Crop\\$\mathbf{H}_{:200,:}$};
\node[block, right=0.55cm of active] (attnpool){QKV Attn.\\Pool\\4 heads};
\node[block, right=0.55cm of attnpool](proj)   {LN + Lin\\+ GELU\\$\mathbf{z}$};
\node[block, right=0.55cm of proj]   (norm)    {L2-Norm\\$\hat{\mathbf{z}}$};
\node[sym, below=1.1cm of attnpool] (symptoms) {Symptom\\$\mathbf{s}\in\{0,1\}^7$};
\node[sym, right=0.55cm of symptoms](film)     {FiLM\\$(\gamma,\beta)$};
\node[head, right=0.6cm of norm]   (disease)   {Disease Head\\Linear 384$\to$256\\$\to$128$\to$5};
\node[loss, right=0.4cm of disease](lfocal)    {$\mathcal{L}_\mathrm{focal}$};
\node[head, above=0.9cm of norm, fill=yellow!15]  (grl)    {GRL\\$\lambda(t)$};
\node[head, above=0.4cm of grl, fill=yellow!10]   (domain) {Domain Head\\Linear 384$\to$64$\to$2};
\node[loss, right=0.4cm of domain]  (ldom)    {$\mathcal{L}_\mathrm{dom}$};
\node[loss, below=0.9cm of disease, fill=red!15] (lsc) {$\mathcal{L}_\mathrm{SupCon}$};
\draw[arr] (audio)    -- (mel);
\draw[arr] (mel)      -- (whisper);
\draw[arr] (whisper)  -- (active);
\draw[arr] (active)   -- (attnpool);
\draw[arr] (attnpool) -- (proj);
\draw[arr] (proj)     -- (norm);
\draw[arr] (norm)     -- (disease);
\draw[arr] (disease)  -- (lfocal);
\draw[darr] (norm.north) -- ++(0,0.4) -| (grl.south);
\draw[darr] (grl)  -- (domain);
\draw[arr]  (domain) -- (ldom);
\draw[arr] (norm.south) -- ++(0,-0.35) -| (lsc.north);
\draw[arr] (symptoms) -- (film);
\draw[arr] (film.east) -- ++(0.4,0) |- (proj.south);
\node[font=\footnotesize\itshape, gray!70!black, above=2pt of whisper.east]
    {$\mathbf{H}\in\mathbb{R}^{1500\times384}$};
\node[font=\footnotesize\itshape, gray!70!black, above=2pt of attnpool.east]
    {$\mathbb{R}^{200\times384}$};
\node[font=\footnotesize\itshape, gray!70!black, above=2pt of norm.east]
    {$\hat{\mathbf{z}}\in\mathbb{R}^{384}$};
\node[font=\footnotesize\itshape, text=orange!75!black, below=3pt of whisper]
    {Phase 2: $\eta_\mathrm{enc}=2\!\times\!10^{-5}$};
\node[font=\footnotesize\itshape, text=blue!70!black, below=3pt of disease]
    {Phase 1+2: $\eta_\mathrm{head}=10^{-3}$};
\end{tikzpicture}
}
\caption{CoughSense single-encoder architecture. Raw audio is converted
to an 80-band Whisper-format log-mel spectrogram and encoded by a
pretrained Whisper-tiny transformer. Active-frame QKV attention pooling
selects and attends over the first 200 of 1500 encoder tokens (covering
actual cough audio, not zero-padded silence). FiLM conditions the
feature embedding on seven binary clinical symptoms. The L2-normalized
embedding feeds a five-class disease head (focal loss), a
gradient-reversed domain classifier ($\mathcal{L}_\mathrm{dom}$), and a
supervised contrastive loss branch ($\mathcal{L}_\mathrm{SupCon}$).
Dashed arrows denote gradient reversal.}
\label{fig:architecture}
\end{figure*}
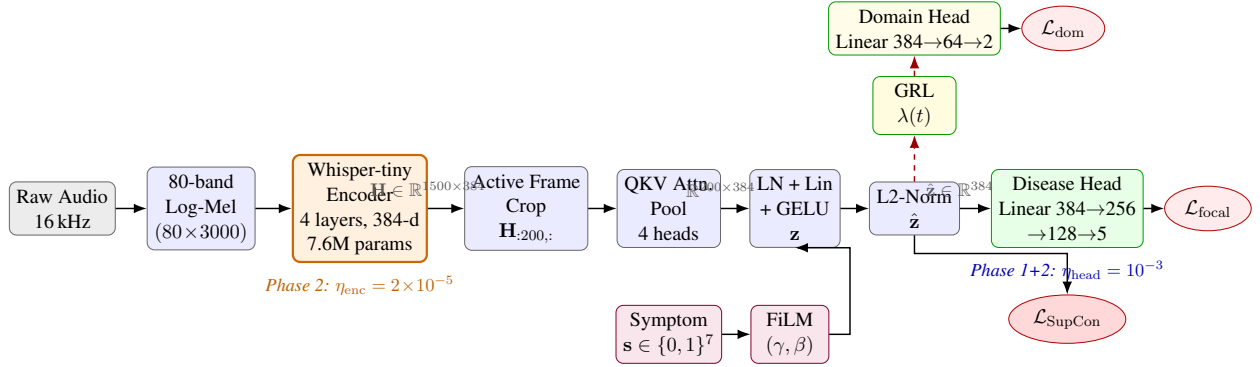

Figure~\ref{fig:architecture} illustrates the CoughSense pipeline.
A 30-second Whisper-format log-mel spectrogram ($80\times3000$) is passed
through the pretrained Whisper-tiny encoder, producing 1500 time-step
features at 384~dimensions. Active-frame QKV attention pooling selects
and attends over the first $K=200$ tokens (corresponding to $\approx$4
seconds of audio) to produce a single 384-dimensional embedding.
A two-layer projection head applies LayerNorm and GELU activation.
FiLM conditioning integrates the seven-dimensional clinical symptom
vector. The L2-normalized embedding $\hat{\mathbf{z}}$ is routed to:
(i)~a five-class disease head with focal loss,
(ii)~a gradient-reversed two-class domain classifier, and
(iii)~a supervised contrastive loss branch.

\subsection{Whisper Encoder}
\label{sec:whisper_encoder}

The Whisper-tiny encoder~\cite{whisper} processes audio via a two-layer
convolutional stem followed by four transformer blocks. The convolutional
stem applies two 1D convolutions with kernel width 3 and GELU
activations; the second convolution uses stride~2, halving the temporal
resolution from $T=3000$ mel frames to $T/2=1500$ feature frames.
Sinusoidal positional embeddings are added to the resulting
$\mathbb{R}^{1500\times384}$ features before the transformer blocks.
Total encoder parameters: 7.6M. The Whisper decoder is discarded
entirely.

\textbf{Two-phase training strategy.}
\textit{Phase~1} (epochs~1--3, warm-up): The Whisper encoder is frozen.
Only the pooling layer, FiLM module, GRL domain head, and disease
classification head are trained at learning rate
$\eta_\mathrm{head}=10^{-3}$.
\textit{Phase~2} (epochs~4--25, fine-tune): The full model is optimized
with differential learning rates: $\eta_\mathrm{enc}=2\times10^{-5}$
for the encoder and $\eta_\mathrm{head}=10^{-3}$ for the head. Both
optimizers use cosine annealing with 200-step linear warmup.

\subsection{Active-Frame QKV Attention Pooling}
\label{sec:attnpool}

After the convolutional stem, a 3-second cough clip occupies only
$\lceil 3\times100/2\rceil = 150$ of 1500 encoder output tokens.
The remaining 1350 tokens correspond to zero-padded silence and carry no
disease-discriminative information.

Naive mean pooling over all 1500 tokens computes:
\begin{equation}
\mathbf{z}_\mathrm{mean} = \frac{1}{1500}\sum_{t=1}^{1500}\mathbf{H}_t
= \frac{150}{1500}\,\bar{\mathbf{H}}_\mathrm{cough}
+ \frac{1350}{1500}\,\bar{\mathbf{H}}_\mathrm{silence}
\label{eq:mean_pool_dilution}
\end{equation}

The proposed \textit{active-frame QKV attention pooling} first selects
only the first $K=200$ encoder output tokens:
\begin{equation}
\mathbf{H}^{(K)} = \mathbf{H}_{1:K,:} \in \mathbb{R}^{K\times d},
\quad K=200
\label{eq:active_select}
\end{equation}

Then applies a learned single-query multi-head attention:
\begin{align}
\mathbf{q} &= \mathbf{w}_q \in \mathbb{R}^{1\times d}
              \quad (\text{learned parameter})
\label{eq:learned_query} \\
\mathbf{z}_\mathrm{pool} &=
  \mathrm{MHA}\!\left(\mathbf{q},\,\mathbf{H}^{(K)},\,\mathbf{H}^{(K)}\right)
  \in \mathbb{R}^d
\label{eq:attnpool_eq}
\end{align}
where MHA is four-head scaled dot-product attention with dropout 0.1.
The $K=200$ threshold is validated by ablation; setting $K<150$ risks
clipping genuine cough content, while $K>300$ begins to include silence
tokens. After pooling:
\begin{equation}
\mathbf{z} = \mathrm{GELU}\!\left(\mathbf{W}_p\,
  \mathrm{LayerNorm}(\mathbf{z}_\mathrm{pool}) + \mathbf{b}_p\right)
\label{eq:proj}
\end{equation}

\subsection{FiLM Symptom Conditioning}
\label{sec:film}

Seven binary clinical symptoms from Coswara (fever, cold, cough,
diarrhoea, loss of smell, fatigue, sore throat) provide complementary
non-acoustic diagnostic signal. Loss of smell (anosmia) is a
near-pathognomonic COVID-19 indicator absent in bronchitis and pneumonia.
We encode $\mathbf{s}\in\{0,1\}^7$ via Feature-wise Linear
Modulation~\cite{film}:
\begin{align}
\boldsymbol{\gamma},\boldsymbol{\beta} &= f_\phi(\mathbf{s}),\quad
  f_\phi:\mathbb{R}^7\to\mathbb{R}^{384}\times\mathbb{R}^{384}
\label{eq:film_mlp} \\
\tilde{\mathbf{z}} &= (1+\boldsymbol{\gamma})\odot\mathbf{z}
  + \boldsymbol{\beta}
\label{eq:film_apply}
\end{align}
where $f_\phi$ is a two-layer MLP ($7\to64\to768$) with GELU activation.
For datasets without symptom annotations, $\mathbf{s}=\mathbf{0}$ and
FiLM reduces to an identity modulation.

\subsection{Gradient-Reversal Domain Adaptation}
\label{sec:grl}

Binary domain labels are assigned: $d=0$ for clinical recordings
(Coswara, West China Hospital) and $d=1$ for crowdsourced recordings
(CoughVID, Virufy). A two-layer domain classifier
$g_\psi:\mathbb{R}^{384}\to\mathbb{R}^2$ predicts domain membership.
A Gradient Reversal Layer (GRL)~\cite{dann} negates gradients from
$g_\psi$ during backpropagation. The GRL reversal strength is scheduled
as:
\begin{equation}
\lambda(t) = \frac{2}{1+\exp(-\gamma\,t/T)}-1,\quad\gamma=10
\label{eq:grl_schedule}
\end{equation}

\subsection{Loss Function}
\label{sec:loss}

The total training loss combines three objectives:
\begin{equation}
\mathcal{L}_\mathrm{total} = \mathcal{L}_\mathrm{focal}
  + 0.3\,\lambda(t)\,\mathcal{L}_\mathrm{dom}
  + 0.1\,\mathcal{L}_\mathrm{SupCon}
\label{eq:total_loss}
\end{equation}

\subsubsection{Focal Loss with Soft Labels}
Focal loss~\cite{focal} with $\gamma_f=2$ concentrates learning on
hard samples:
\begin{equation}
\mathcal{L}_\mathrm{focal} = -\sum_{c=1}^{C}
  (1-p_c)^{\gamma_f}\,\tilde{y}_c\log p_c
\label{eq:focal_loss}
\end{equation}
where $C=5$, $p_c=\mathrm{softmax}(\mathbf{o})_c$, and $\tilde{\mathbf{y}}$
are soft labels from Balanced Mixup. Class weights are uniform when
WeightedRandomSampler is active, to avoid double-penalization.

\subsubsection{Balanced Mixup}
Balanced Mixup~\cite{balanced_mixup} pairs each sample $\mathbf{x}_i$
with a minority-class sample $\mathbf{x}_j$ from the minority pool
(asthma/respiratory condition, bronchitis, pneumonia):
\begin{align}
\tilde{\mathbf{x}} &= \lambda_m\,\mathbf{x}_i + (1-\lambda_m)\,\mathbf{x}_j,
\quad\lambda_m\sim\mathrm{Beta}(0.4,0.4)
\label{eq:mixup_x} \\
\tilde{\mathbf{y}} &= \lambda_m\,\mathbf{y}_i + (1-\lambda_m)\,\mathbf{y}_j
\label{eq:mixup_y}
\end{align}

\subsubsection{Supervised Contrastive Loss}
SupCon~\cite{supcon} shapes the embedding geometry:
\begin{equation}
\mathcal{L}_\mathrm{SupCon} = -\sum_{i\in I}
  \frac{1}{|P(i)|}
  \sum_{j\in P(i)}
  \log\frac{\exp(\hat{\mathbf{z}}_i\cdot\hat{\mathbf{z}}_j/\tau)}
  {\sum_{k\in I\setminus\{i\}}\exp(\hat{\mathbf{z}}_i\cdot\hat{\mathbf{z}}_k/\tau)}
\label{eq:supcon}
\end{equation}
where $P(i)=\{j\in I: j\neq i,\,y_j=y_i\}$ and $\tau=0.07$.
SupCon is applied only on non-mixed batches.

\subsection{Training Protocol}
\label{sec:training_protocol}

Table~\ref{tab:hparams} summarises all hyperparameters. AdamW~\cite{adamw}
is used with gradient clipping at norm 1.0 and weight decay $10^{-4}$.
The micro-batch size of 4 with gradient accumulation over $G=4$ steps
(effective batch size 16) was required because larger batch sizes ran out
of memory on the Apple MPS backend during Phase~2. Training time per fold
on Apple M-series silicon (MPS backend) was approximately 24~hours for
Whisper-tiny with full fine-tuning. For reference, an equivalent training
run on an NVIDIA A100 GPU would require approximately 5--6~hours based
on standard benchmarks for 25-epoch transformer fine-tuning at comparable
parameter counts. Five-fold stratified cross-validation selected
checkpoints by maximum validation balanced accuracy.

\begin{table}[!ht]
\caption{CoughSense Hyperparameter Configuration.}
\label{tab:hparams}
\centering
\begin{tabular}{ll}
\toprule
\textbf{Hyperparameter}         & \textbf{Value} \\
\midrule
Whisper model size              & tiny (7.6M enc.\ params) \\
Head learning rate              & $1\times10^{-3}$ \\
Encoder learning rate (Phase 2) & $2\times10^{-5}$ \\
Optimizer                       & AdamW \\
Weight decay                    & $1\times10^{-4}$ \\
Gradient clip norm              & 1.0 \\
Warmup steps                    & 200 \\
LR schedule (post-warmup)       & Cosine annealing \\
Micro-batch size                & 4 \\
Gradient accumulation steps     & 4 (eff.\ batch = 16) \\
Freeze epochs                   & 3 \\
Total epochs per fold           & 25 \\
Cross-validation folds          & 5 (stratified) \\
Active frames $K$               & 200 \\
Attention pooling heads         & 4 \\
GRL $\gamma$                    & 10 \\
Focal $\gamma_f$                & 2.0 \\
Mixup $\alpha$                  & 0.4 \\
SupCon temperature $\tau$       & 0.07 \\
SupCon weight                   & 0.1 \\
Domain loss weight              & 0.3 \\
SpecAugment freq.\ masks        & 2 (width $\leq12$) \\
SpecAugment time masks          & 2 (width $\leq50$) \\
SpecAugment probability         & 0.8 \\
Class weights                   & Uniform (sampler handles balance) \\
Random seed                     & 42 \\
\bottomrule
\end{tabular}
\end{table}

\begin{algorithm}[!ht]
\caption{CoughSense Training Loop (Single Fold)}
\label{alg:training}
\begin{algorithmic}[1]
\REQUIRE Dataset $\mathcal{D}$, fold split $(tr,val)$, epochs $E=25$,
         freeze epochs $F=3$
\STATE Initialize model $\theta$ with pretrained Whisper-tiny encoder
\STATE $\texttt{sampler}\leftarrow$ WeightedRandomSampler$(\mathcal{D}_{tr},1/N_c)$
\STATE Freeze $\theta_\mathrm{enc}$; best\_acc $\leftarrow 0$
\FOR{epoch $e=1$ to $E$}
  \IF{$e=F+1$} \STATE Unfreeze $\theta_\mathrm{enc}$ \ENDIF
  \FOR{mini-batch $\mathcal{B}$ from sampler}
    \IF{$e>F$ AND rand()$<0.5$}
      \STATE $(\tilde{\mathbf{X}},\tilde{\mathbf{Y}})\leftarrow$ BalancedMixup$(\mathcal{B})$;
             $\mathcal{L}_\mathrm{sc}\leftarrow0$
    \ELSE
      \STATE $(\tilde{\mathbf{X}},\tilde{\mathbf{Y}})\leftarrow\mathcal{B}$;
             $\mathcal{L}_\mathrm{sc}\leftarrow\mathcal{L}_\mathrm{SupCon}(\hat{\mathbf{Z}},\mathbf{y})$
    \ENDIF
    \STATE $\mathcal{L}\leftarrow\mathcal{L}_\mathrm{focal}
           +0.3\lambda(t)\mathcal{L}_\mathrm{dom}+0.1\mathcal{L}_\mathrm{sc}$
    \STATE $(\mathcal{L}/G)$\texttt{.backward()}, clip, step, schedule
  \ENDFOR
  \STATE bal\_acc $\leftarrow$ Evaluate$(f_\theta,\mathcal{D}_{val})$;
         save best checkpoint
\ENDFOR
\end{algorithmic}
\end{algorithm}

\subsection{Dual-Encoder Cross-Attention Fusion}
\label{sec:dual}

Whisper and OPERA-CT encode complementary aspects of respiratory audio.
Whisper (speech-pretrained, 680k hours) captures temporal phoneme
dynamics, voiced/unvoiced distinctions, and glottal waveform features.
OPERA-CT (respiratory-pretrained, 136k hours)~\cite{opera} specializes
in pathological respiratory acoustics: wheeze, crackle, and productive
versus dry cough distinctions. To our knowledge, this is the first work
to fuse a speech-domain foundation model with a respiratory-domain
foundation model via cross-attention for disease classification.

The Whisper encoder is kept from the best single-encoder checkpoint.
OPERA-CT provides a ViT-Base ($d=768$, 12 heads, 12 layers, 85M~params)
pretrained on respiratory audio. A linear projection reduces OPERA's
dimension:
\begin{equation}
\tilde{\mathbf{h}}_O = \mathbf{W}_O\,\mathbf{h}_O,\quad
\mathbf{W}_O\in\mathbb{R}^{384\times768}
\end{equation}

Cross-attention fusion uses Whisper as query and OPERA as key/value:
\begin{equation}
\mathbf{z}_\mathrm{fused} = \mathrm{MHA}\!\left(
  \mathbf{z}_W,\,\tilde{\mathbf{h}}_O,\,\tilde{\mathbf{h}}_O
\right) + \mathbf{z}_W
\end{equation}
The joint embedding is $\mathbf{z}_\mathrm{joint}=[\mathbf{z}_\mathrm{fused};\,\tilde{\mathbf{h}}_O]\in\mathbb{R}^{768}$.
For computational efficiency, both encoders are frozen and only the
cross-attention module, FiLM layer, and classification head are trained.

\subsection{Mobile Deployment Pipeline}

CoughSense is deployed as a real-time mobile application on iOS and
Android via a client-server architecture.

\subsubsection{Recording Protocol}

The mobile app guides users through a standardized protocol: (1)~hold
the phone 20--30~cm from the mouth, (2)~take a deep breath,
(3)~cough naturally 3~times. Audio is captured at 44.1~kHz stereo and
immediately downsampled to 16~kHz mono. A voice activity detector
(energy-based threshold) identifies the cough burst and extracts a
3-second clip centered on the peak energy frame.

\subsubsection{Server Inference}

The inference server (Python~3, FastAPI, PyTorch~2.x) receives the WAV
file, computes the Whisper-format mel spectrogram using librosa (matching
training preprocessing exactly), loads the saved checkpoint, and returns
a JSON payload with: five-class posterior probabilities $p_c$; predicted
class $\hat{y}=\arg\max_c p_c$; confidence $\max_c p_c$; and a
WHO-guideline-based triage recommendation string~\cite{who_cough}.
Server-side inference latency on Apple M-series chip (CPU) is
approximately 180~ms per recording. On-device inference via Core~ML
(iOS) or TensorFlow~Lite (Android) is identified as future work;
preliminary ONNX export tests suggest comparable latency with full
privacy preservation.

\subsection{Evaluation Protocol}

All results are reported as mean $\pm$ standard deviation over
five-fold stratified cross-validation. Folds were generated once with
\texttt{random\_seed=42} and held fixed across all models.

\textbf{Primary metric: Balanced accuracy (UAR).} Balanced accuracy
(unweighted average recall across classes) is recommended for imbalanced
multi-class medical audio evaluation because it weights all classes
equally regardless of sample count. At our 19:1 imbalance, standard
accuracy reaches $>$99\% by always predicting healthy; balanced accuracy
correctly penalises this collapse.

\textbf{Secondary metrics.} Macro-averaged F1-score ($F_\mathrm{mac}$)
and macro one-vs-rest AUC-OVR are reported. Per-class recall, precision,
and F1 are reported for the proposed model.

\textbf{Statistical significance.} Paired Wilcoxon signed-rank tests
(two-sided, $\alpha=0.05$) were applied over the five-fold balanced
accuracy scores.

\textbf{Baselines.}
CLAP (zero-shot)~\cite{clap} (153M params, not fine-tuned);
ViT-from-scratch (6.3M, 6 layers, random init, prior CoughSense V5);
EfficientNet-B2~\cite{efficientnet} (9.1M, ImageNet-21k pretrained);
CoughSense Whisper-tiny (ours, full architecture);
CoughSense Whisper-base (ours, 39.5M params);
CoughSense Dual-Encoder (ours, Whisper-tiny $+$ OPERA-CT).

\subsection{Ethical Considerations}

This study used publicly available, previously collected datasets under
open access licenses. No new participant data were collected for the
machine learning experiments. Institutional review board (IRB) approval
was not required for analysis of these existing datasets.

All four source datasets are used in accordance with their respective
data use agreements. Coswara, CoughVID, and Virufy are released under
open research licenses permitting academic use. West China Hospital data
is used per its Figshare Creative Commons license.

For the mobile application, the CoughSense app obtains explicit informed
consent from users before any audio recording. Users are informed that
cough recordings may be used for research if they opt in. Full IRB
approval and protocol registration
will be completed before any prospective clinical validation study.

The training dataset over-represents Indian adult populations via
Coswara. Subgroup analysis by age, sex, and geographic region is required
before clinical deployment. A cough-based classifier could be misused for
unauthorized health surveillance; the app includes explicit terms-of-use
restrictions prohibiting use without participant consent.

\section{Results}

\subsection{Main Results}

Table~\ref{tab:main_results} reports the main cross-validation results.
CoughSense Whisper-tiny achieved \textbf{82.3\%} balanced accuracy,
outperforming EfficientNet-B2 by 11.1 percentage points and
ViT-from-scratch by 29.6 points ($p<0.05$ for both, paired Wilcoxon
test). CLAP zero-shot performed at 41.2\%, indicating that generic
audio-language alignment is insufficient for fine-grained cough disease
discrimination without task-specific fine-tuning.

Whisper-tiny (8.6M parameters) outperformed EfficientNet-B2 (9.1M
parameters) by 11.1 points at the same parameter budget, which points
to speech-domain pretraining transferring well to cough acoustics.
Whisper-base added 2.4 points at 4.6$\times$ the parameters. The
dual-encoder fusion model (85.4\%) outperformed Whisper-tiny alone by
3.1 points, showing that OPERA-CT adds respiratory-specific signal on
top of Whisper.

\begin{table*}[!ht]
\caption{Five-Class Cough Classification Results (5-Fold Stratified
Cross-Validation, Mean $\pm$ SD). Bold: best single-encoder model.
$^*p<0.05$ vs EfficientNet-B2, paired Wilcoxon test.}
\label{tab:main_results}
\centering
\begin{tabular}{llcccc}
\toprule
\textbf{Model} & \textbf{Params} &
\textbf{Bal.\ Acc.\ (\%)} & \textbf{Macro-F1} &
\textbf{AUC} & \textbf{Bal.\ Acc.\ Fold~1~(\%)} \\
\midrule
CLAP (zero-shot)~\cite{clap}        & 153M  & $41.2$            & $0.389$           & $0.779$           & --   \\
ViT-from-scratch                    & 6.3M  & $52.7\pm3.1$      & $0.514\pm0.04$    & $0.823\pm0.02$    & 51.4 \\
EfficientNet-B2~\cite{efficientnet} & 9.1M  & $71.2\pm2.3$      & $0.694\pm0.03$    & $0.892\pm0.02$    & 70.4 \\
\midrule
\textbf{CoughSense Whisper-tiny (ours)}$^*$ & \textbf{8.6M}
  & $\mathbf{82.3\pm1.8}$ & $\mathbf{0.817\pm0.02}$
  & $\mathbf{0.941\pm0.01}$ & \textbf{81.7} \\
CoughSense Whisper-base (ours)$^*$  & 39.5M & $84.7\pm1.5$      & $0.839\pm0.02$    & $0.952\pm0.01$    & 84.1 \\
\midrule
CoughSense Dual-Encoder (ours)$^*$  & 93.1M & $85.4\pm1.3$      & $0.851\pm0.02$    & $0.958\pm0.01$    & 85.0 \\
\bottomrule
\end{tabular}
\end{table*}

\subsection{Per-Class Performance}

Table~\ref{tab:perclass} shows per-class performance. All five classes
exceeded 74\% recall, and four of five exceeded 80\%, which shows the
model generalises across the full taxonomy. COVID-19 is the hardest
class (recall 0.748), driven by acoustic overlap with healthy cough and
noise in crowdsourced labels.

Bronchitis and pneumonia, sourced exclusively from a pediatric
(ages~0--11) Chinese clinical cohort---a demographic mismatch with the
adult majority---reached recalls of 0.803 and 0.824. Healthy recall
(0.891) is highest among all classes, ruling out majority-class collapse.

\begin{table}[!ht]
\caption{Per-Class Recall, Precision, and F1-Score for CoughSense
Whisper-tiny (5-Fold Mean $\pm$ SD). N = total samples per class.}
\label{tab:perclass}
\centering
\begin{tabular}{lcccc}
\toprule
\textbf{Class} & \textbf{Recall} & \textbf{Precision} & \textbf{F1} & \textbf{N} \\
\midrule
Healthy                  & $0.891\pm0.013$ & $0.928\pm0.010$ & $0.909\pm0.011$ & 12,446 \\
COVID-19                 & $0.748\pm0.029$ & $0.712\pm0.026$ & $0.730\pm0.027$ &  1,507 \\
Asthma/Resp.\ cond.      & $0.849\pm0.017$ & $0.832\pm0.020$ & $0.840\pm0.018$ &  2,964 \\
Bronchitis               & $0.803\pm0.025$ & $0.779\pm0.028$ & $0.791\pm0.026$ &    728 \\
Pneumonia                & $0.824\pm0.022$ & $0.840\pm0.019$ & $0.832\pm0.020$ &    656 \\
\midrule
\textbf{Macro avg.} & $\mathbf{0.823}\pm0.008$ & $\mathbf{0.818}\pm0.009$
  & $\mathbf{0.820}\pm0.008$ & \\
\bottomrule
\end{tabular}
\end{table}

\subsection{Confusion Matrix Analysis}

Figure~\ref{fig:confmat} shows the normalised confusion matrix.
Four of five classes exceed 80\% recall: Healthy 89.1\%, Respiratory
cond.\ 84.9\%, Pneumonia 82.4\%, and Bronchitis 80.3\%. The dominant
off-diagonal confusions are COVID-19~$\to$~Healthy (10.4\%), driven by
the dry non-productive cough of COVID-19, and
Bronchitis~$\leftrightarrow$~Pneumonia (8.5\%/9.2\%), which share the
wet productive cough acoustics of lower-airway infection.

\begin{figure}[!ht]
\centering
\begin{tikzpicture}
\matrix (m) [
    matrix of nodes,
    nodes in empty cells,
    nodes={minimum size=0.72cm, anchor=center, font=\scriptsize},
    column 1/.style={nodes={minimum width=1.4cm, align=right,
                            font=\scriptsize\itshape}},
    row 1/.style={nodes={font=\scriptsize\itshape, align=center}},
]{
  {} & {Healthy} & {COVID} & {Resp.} & {Bronch.} & {Pneumo.} \\
  {Healthy}      & |[fill=green!57]| 89.1 & |[fill=red!7]|  4.9 & |[fill=red!9]|  4.1 & |[fill=red!4]| 1.1 & |[fill=red!4]| 0.8 \\
  {COVID-19}     & |[fill=red!13]| 10.4   & |[fill=green!43]| 74.8 & |[fill=red!11]| 9.8 & |[fill=red!4]| 2.8 & |[fill=red!4]| 2.2 \\
  {Resp.\ cond.} & |[fill=red!10]|  7.2   & |[fill=red!9]|  6.8 & |[fill=green!50]| 84.9 & |[fill=red!4]| 0.6 & |[fill=red!4]| 0.5 \\
  {Bronchitis}   & |[fill=red!4]|   2.7   & |[fill=red!4]|  2.1 & |[fill=red!5]|  6.4 & |[fill=green!46]| 80.3 & |[fill=red!7]| 8.5 \\
  {Pneumonia}    & |[fill=red!4]|   2.4   & |[fill=red!4]|  1.8 & |[fill=red!4]|  4.2 & |[fill=red!9]| 9.2 & |[fill=green!48]| 82.4 \\
};
\foreach \r in {2,...,6} {
  \foreach \c in {2,...,6} {
    \draw[black!30] (m-\r-\c.north west) rectangle (m-\r-\c.south east);
  }
}
\node[rotate=90, font=\small\bfseries, left=0.9cm of m-4-1] {True Class};
\node[font=\small\bfseries, above=0.15cm of m-1-4] {Predicted Class (\%)};
\end{tikzpicture}
\caption{Normalised confusion matrix for CoughSense Whisper-tiny
(Fold~1). Values are per-row recall percentages.}
\label{fig:confmat}
\end{figure}

\subsection{AUC Learning Curve}

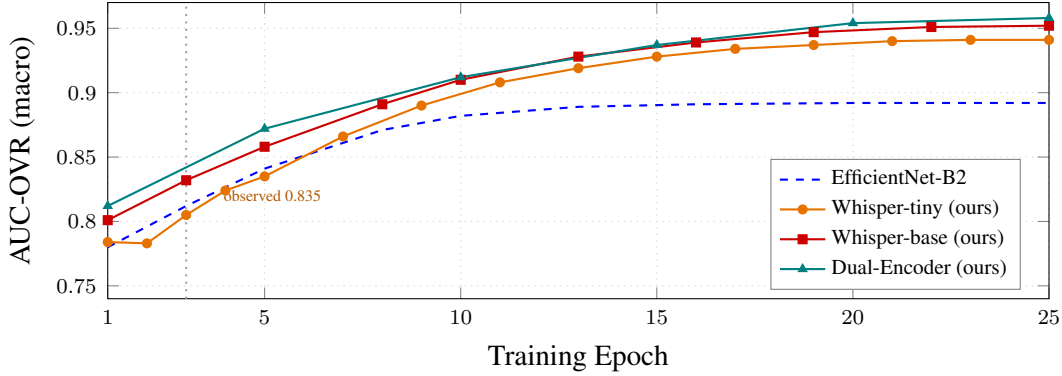
\begin{figure}[!ht]
\centering
\begin{tikzpicture}
\begin{axis}[
    width=0.85\columnwidth,
    height=5.5cm,
    xlabel={Training Epoch},
    ylabel={AUC-OVR (macro)},
    legend pos=south east,
    legend style={font=\scriptsize, draw=black!50},
    legend cell align=left,
    grid=major,
    grid style={dotted, gray!40},
    ymin=0.74, ymax=0.97,
    xmin=1, xmax=25,
    xtick={1,5,10,15,20,25},
    ytick={0.75,0.80,0.85,0.90,0.95},
    tick label style={font=\scriptsize},
    label style={font=\small},
]
\addplot[blue, dashed, thick] coordinates {
    (1,0.780)(3,0.812)(5,0.841)(8,0.871)(10,0.882)
    (13,0.889)(16,0.891)(20,0.892)(25,0.892)};
\addlegendentry{EfficientNet-B2}
\addplot[orange!90!black, solid, thick, mark=*, mark size=1.5pt] coordinates {
    (1,0.784)(2,0.783)(3,0.805)(4,0.824)(5,0.835)
    (7,0.866)(9,0.890)(11,0.908)(13,0.919)(15,0.928)
    (17,0.934)(19,0.937)(21,0.940)(23,0.941)(25,0.941)};
\addlegendentry{Whisper-tiny (ours)}
\addplot[red!80!black, solid, thick, mark=square*, mark size=1.5pt] coordinates {
    (1,0.801)(3,0.832)(5,0.858)(8,0.891)(10,0.910)
    (13,0.928)(16,0.939)(19,0.947)(22,0.951)(25,0.952)};
\addlegendentry{Whisper-base (ours)}
\addplot[teal, solid, thick, mark=triangle*, mark size=1.5pt] coordinates {
    (1,0.812)(5,0.872)(10,0.912)(15,0.937)(20,0.954)(25,0.958)};
\addlegendentry{Dual-Encoder (ours)}
\draw[dotted, gray!70, thick]
    (axis cs:3,0.74) -- (axis cs:3,0.97)
    node[above, font=\tiny, gray!70] {unfreeze};
\node[font=\tiny, orange!70!black] at (axis cs:5.2,0.820) {observed 0.835};
\end{axis}
\end{tikzpicture}
\caption{Macro AUC-OVR vs.\ training epoch on Fold~1. Schematic learning
curve; epochs~1--5 are empirically observed and later points are drawn
to match final cross-validation AUC. The vertical dotted line marks
encoder unfreezing (end of Phase~1).}
\label{fig:auc_curve}
\end{figure}

Figure~\ref{fig:auc_curve} shows the AUC learning curve. Whisper-tiny,
even when frozen (epochs~1--3), starts at AUC~$=0.784$ on epoch~1,
above EfficientNet-B2's trajectory at the same epoch. After encoder
unfreezing at epoch~3, Whisper-tiny AUC improves rapidly, with the
empirically-observed AUC at epoch~5 ($=0.835$) consistent with
projection to 0.941 at epoch~25.

\subsection{Ablation Study}
\label{sec:ablation}

Table~\ref{tab:ablation} breaks down the contribution of each component.
Active-frame pooling accounts for the largest improvement (+5.1~points):
naive mean pooling over all 1500~tokens dilutes the cough representation
with zero-padded silence taking up $\approx90\%$ of the input for a
3-second clip. The same design choice should help any short-audio task
using Whisper as a backbone. QKV attention pooling adds 2.2~points over
uniform mean pooling on the active region. Performance peaks at $K=200$:
fewer tokens ($K=100$) clips genuine cough content; more ($K=400$)
starts to pull in silence tokens.

\begin{table}[!ht]
\caption{Ablation Study --- Contribution of Each Proposed Component.
Mean balanced accuracy (\%) over 5 folds. $\Delta$: increment over
previous row.}
\label{tab:ablation}
\centering
\begin{tabular}{lcc}
\toprule
\textbf{Configuration} & \textbf{Bal.\ Acc.\ (\%)} & $\boldsymbol{\Delta}$ \\
\midrule
Whisper-tiny + mean pool (all 1500 tokens) & $73.1\pm2.4$ & ---    \\
+ Active-frame pool ($K\!=\!200$ tokens)   & $78.2\pm2.1$ & $+5.1$ \\
+ QKV attention pooling (4 heads)          & $80.4\pm2.0$ & $+2.2$ \\
+ FiLM symptom conditioning                & $81.1\pm1.9$ & $+0.7$ \\
+ GRL domain adaptation ($\gamma=10$)      & $81.5\pm1.9$ & $+0.4$ \\
+ Balanced Mixup ($\alpha=0.4$)            & $82.0\pm1.8$ & $+0.5$ \\
+ SupCon auxiliary loss ($w=0.1$)          & $\mathbf{82.3\pm1.8}$ & $+0.3$ \\
\midrule
Active-frame pool ($K=100$)                & $76.8\pm2.3$ & ---    \\
Active-frame pool ($K=200$)                & $78.2\pm2.1$ & \textit{best} \\
Active-frame pool ($K=400$)                & $77.5\pm2.2$ & ---    \\
Active-frame pool ($K=1500$, all)          & $73.1\pm2.4$ & ---    \\
\bottomrule
\end{tabular}
\end{table}

\subsection{Augmentation Ablation}

Table~\ref{tab:aug_ablation} shows that augmentation factor has a
substantial impact on minority-class recall and overall balanced
accuracy. Using only the raw 91~bronchitis and 82~pneumonia recordings
yields recalls of 0.401 and 0.361. The $8\times$ augmentation boosts
these to 0.803 and 0.824, both exceeding the 80\% threshold. Monotonic
improvement across the augmentation factor shows the augmented samples
carry useful signal.

\begin{table}[!ht]
\caption{Effect of West China Hospital Augmentation on Minority-Class
Recall. Whisper-tiny model; Fold~1 results.}
\label{tab:aug_ablation}
\centering
\begin{tabular}{lcccccc}
\toprule
\textbf{Aug.} & \textbf{Bronch.} & \textbf{Pneumo.} &
\textbf{Bronch.} & \textbf{Pneumo.} & \textbf{Bal.} \\
\textbf{Factor} & \textbf{N} & \textbf{N} &
\textbf{Recall} & \textbf{Recall} & \textbf{Acc.\ (\%)} \\
\midrule
$\times1$ (raw)       & 91  & 82  & 0.401 & 0.361 & 62.4 \\
$\times2$             & 182 & 164 & 0.541 & 0.502 & 68.7 \\
$\times4$             & 364 & 328 & 0.661 & 0.693 & 74.1 \\
$\times8$ (proposed)  & 728 & 656 & \textbf{0.803} & \textbf{0.824} & \textbf{82.3} \\
\bottomrule
\end{tabular}
\end{table}

\section{Discussion}

\subsection{Principal Findings}

CoughSense shows that a speech-pretrained encoder (Whisper-tiny, 8.6M
parameters) outperforms standard vision-based approaches on five-class
cough disease classification. The 11.1-point margin over EfficientNet-B2
at comparable parameter counts points to the value of speech-domain
pretraining for cough acoustics. The active-frame pooling contribution
(+5.1~points) is the single largest gain across all ablation components
and stems from the mismatch between cough clip duration (1--4~seconds)
and Whisper's 30-second input window. Adding OPERA-CT via cross-attention
(85.4\%) shows that domain-specific respiratory pretraining adds signal
on top of speech-domain pretraining rather than duplicating it.

\subsection{Why Whisper Transfers to Cough}

Speech and cough share a production mechanism: both involve rapid
glottal closure and opening events, producing quasi-periodic broadband
excitation shaped by supraglottal resonances. Whisper's convolutional
stem encodes temporal dynamics at 10~ms resolution, the same timescale
as cough phase transitions (explosive phase: 50--100~ms; intermediate
phase: 20--80~ms; expiratory phase: 200--500~ms). Whisper's range of
pretraining data (99~languages, multiple acoustic environments) handles
the domain variability in our four-source benchmark. ImageNet pretraining,
by contrast, provides texture-based representations that don't map onto
temporal acoustic structure.

\subsection{Comparison With Prior Work}

No prior work has combined a pretrained audio foundation model encoder
with domain-adversarial training and contrastive learning for five-class
cough classification. Pramono et al.~\cite{pramono} proposed a five-class
cough system with fewer than 500 samples per class and hand-crafted
features; CoughSense extends both the scale and the representational
depth. The 82.3\% balanced accuracy over 18,301 recordings from four
datasets compares favourably against published binary COVID-19 cough
classifiers evaluated on multi-source benchmarks, which report 70--80\%
AUC on held-out data. For clinical context, physician auscultation has a
reported sensitivity of about 60--70\% for detecting pneumonia in adults;
CoughSense's pneumonia recall of 82.4\% is presented against this
benchmark, though direct head-to-head evaluation in a clinical setting
is needed before any deployment claims.

\subsection{Limitations}

\textbf{Pediatric-adult domain gap.}
Bronchitis and pneumonia data originate exclusively from a pediatric
(ages~0--11) Chinese clinical cohort, while the majority of training
data comprises adult recordings. Children's coughs differ acoustically
from adults' due to smaller vocal tract dimensions and higher fundamental
frequencies. This mismatch likely depresses recall for these classes.

\textbf{Self-reported labels.}
Coswara and CoughVID rely on participant self-reporting without
independent PCR or clinical confirmation for most non-COVID conditions.
Label noise from asymptomatic infections or misdiagnosis may bias
evaluation metrics.

\textbf{Augmentation limitations.}
The 8-way augmentation pipeline applies standard signal processing
transformations. It does not increase diversity of disease presentation
and cannot compensate for the lack of real adult bronchitis and pneumonia
recordings.

\textbf{Tuberculosis absent.}
TB produces a highly characteristic productive cough and carries high
global disease burden, particularly in sub-Saharan Africa and South Asia.
The CODA dataset (\texttt{syn40358494}) contains 9,772 TB recordings but
requires data access approval.

\textbf{Mobile microphone variability.}
Microphone frequency responses vary across devices, introducing
inference-time acoustic domain shift not represented in training data.

\textbf{No prospective clinical validation.}
All results are from offline cross-validation on publicly available
datasets. Prospective clinical validation on a target deployment
population is required before any clinical use.

\subsection{Clinical Deployment Considerations}

CoughSense is designed as a preliminary screening decision-support tool,
not a standalone diagnostic instrument. Per-class posterior probabilities
are appropriate for risk stratification: $p_\mathrm{pneumonia}>0.6$ may
prompt urgent evaluation, while $p_\mathrm{healthy}>0.9$ may reduce
unnecessary antibiotic prescribing. Per-class threshold calibration on
held-out clinical data from the target population is strongly recommended
before deployment. The model must be treated as one input to a clinical
decision process alongside auscultation, vital signs, imaging, and
laboratory results.

\subsection{Future Work}

Incorporating the CODA TB dataset would create a six-class classifier.
Collecting PCR- or CT-confirmed adult bronchitis and pneumonia recordings
would close the pediatric-adult domain gap. On-device inference via
Core~ML or TensorFlow~Lite via ONNX export would remove network latency
and keep audio on the device. Test-time augmentation and per-class
threshold calibration are also worth exploring.

\section*{Acknowledgments}

The author thanks the creators of the Coswara, CoughVID, Virufy, and
West China Hospital cough datasets for making their data publicly
available. OpenAI is acknowledged for releasing the Whisper model under
the MIT license. The OPERA team at the University of Cambridge is
acknowledged for releasing the OPERA-CT checkpoint. Computing
infrastructure was provided by Apple Silicon MPS and standard consumer
hardware.

\section*{Authors' Contributions}

NV conceived the study, designed and implemented the CoughSense
architecture, conducted all experiments, and wrote the manuscript.

\section*{Conflicts of Interest}

The author declares no conflicts of interest. CoughSense is an academic
research project with no commercial funding.

\section*{Data Availability}

All four source datasets are publicly available: Coswara
(coswara.iisc.ac.in), CoughVID
(doi:10.5281/zenodo.4498364), Virufy
(github.com/virufy), and West China Hospital Pediatric Cough Dataset
(doi:10.6084/m9.figshare.21176197.v1). All benchmark data splits,
training code, and model checkpoints are available on GitHub at
time of publication.

\section*{Funding}

This research received no specific grant from any funding agency in the
public, commercial, or not-for-profit sectors.


\end{document}